\documentclass[conference]{IEEEtran} 
\IEEEoverridecommandlockouts
\usepackage{cite}
\usepackage{amsmath,amssymb,amsfonts}
\usepackage{algorithmic}
\usepackage{graphicx}
\usepackage{textcomp}
\usepackage{xcolor}
\usepackage{xurl}
\usepackage{float}
\def\BibTeX{{\rm B\kern-.05em{\sc i\kern-.025em b}\kern-.08em
    T\kern-.1667em\lower.7ex\hbox{E}\kern-.125emX}}
\usepackage{balance}

\begin{document}

\title{Product-oriented Product--Process--Resource Asset Network
and its Representation in AutomationML for Asset Administration Shell}

\author{
    \IEEEauthorblockN{
        Sára Strakošová\IEEEauthorrefmark{1}\IEEEauthorrefmark{3},
        Petr Novák\IEEEauthorrefmark{3},
        Petr Kadera\IEEEauthorrefmark{3}
    }
    \IEEEauthorblockA{
        \IEEEauthorrefmark{1}\textit{Faculty of Mechanical Engineering, Czech Technical University in Prague}, Prague, Czech Republic\\
        sara.strakosova@fs.cvut.cz \\
        \IEEEauthorrefmark{3}\textit{Czech Institute of Informatics, Robotics and Cybernetics, Czech Technical University in Prague}, Prague, Czech Republic\\
        petr.novak@cvut.cz, petr.kadera@cvut.cz
    }
}

\newcommand\textline[4][t]{
  \par\smallskip\noindent\parbox[#1]{1\textwidth}{\raggedright#2}%
  \parbox[#1]{.2\textwidth}{\centering#3}%
  \parbox[#1]{.4\textwidth}{\raggedleft\texttt{#4}}\par\smallskip%
}

\makeatletter

\def\ps@IEEEtitlepagestyle{%
  \def\@oddfoot{\mycopyrightnotice}%
  \def\@evenfoot{}%
}
\def\mycopyrightnotice{%
 \textline[t]{\small \copyright 2024 IEEE.  Personal use of this material is permitted. Permission from IEEE must be obtained for all other uses, in any current or future media, including reprinting/republishing this material for advertising or promotional purposes, creating new collective works, for resale or redistribution to servers or lists, or reuse of any copyrighted component of this work in other works.}{\thepage}{}
  \gdef\mycopyrightnotice{}
}

\maketitle

\begin{abstract}
Current products, especially in the automotive sector, pose complex technical systems having a multi-disciplinary mechatronic nature. Industrial standards supporting system engineering and production typically (i) address the production phase only, but do not cover the complete product life cycle, and (ii) focus on production processes and resources rather than the products themselves. The presented approach is motivated by incorporating the impacts of the end-of-life phase of the product life cycle into the engineering phase. This paper proposes a modeling approach coming up from the Product-Process-Resource (PPR) modeling paradigm. It combines requirements on (i) respecting the product structure as a basis for the model, and (ii) incorporates repairing, remanufacturing, or upcycling within cyber-physical production systems. The proposed model called PoPAN should accompany the product during the entire life cycle as a digital shadow encapsulated within the Asset Administration Shell of a product. To facilitate the adoption of the proposed paradigm, the paper also proposes serialization of the model in the AutomationML data format. The model is demonstrated on a use-case for disassembling electric vehicle batteries to support their remanufacturing for stationary battery applications.
\end{abstract}



\section{Introduction}

The significant environmental impact of manufacturing industries underscores the importance of addressing sustainability in the context of recycling and remanufacturing of products~\cite{EnvImp}. Although the term "sustainability" is increasingly used in marketing,  leading to skepticism and negative perceptions. Distinguishing genuine sustainability efforts from mere marketing strategies is essential. In this paper, we introduce a formalization to model products and production processes, which has the potential to contribute to sustainability specifically at the end of a product life-cycle.

This paper introduces the Product-oriented Product–Process–Resource Asset Network (PoPAN) formalization, which builds upon the Product–Process–Resource (PPR)~\cite{AHMAD201833} and the Product–Process–Resource Asset Network (PAN)~\cite{Biffl2021CBI} approaches. Current modeling approaches (including the original PAN) are frequently focused on production processes. This perspective is substantiated by numerous industrial standards such as ISA-95 or IEC~62264, Asset Administration Shell, and VDI/VDE~3682. On the contrary, the PoPAN spotlights the product structure as a basis and it assigns production processes/operations to product components.

PoPAN offers a structured description of a product, incorporating its components (represented as products), processes, and resources. This approach enables the design of recycling and remanufacturing processes for products that require disassembly for efficient recycling. Furthermore, PoPAN can function as a digital shadow~\cite{DS}, accompanying the product throughout its entire life-cycle and encapsulating all relevant information. It serves as a record of the product's evolution, including any modifications or alterations it undergoes during its journey, such as missing components like screws. 
An additional advantage of employing PoPAN for recycling lies in the data collection aspect. Sustainability initiatives begin at the product design phase, where prioritizing recyclability is paramount. By gathering relevant data from recycling processes, manufacturers can analyze and utilize this information to enhance the recyclability of future product designs. This iterative process leads to more sustainable designs and facilitates easier recycling, contributing to overall environmental sustainability efforts.


To facilitate the adoption of this paradigm, the paper also proposes the serialization of the model in the AutomationML~\cite{Drath2008} data format. This standardized format enhances interoperability and ease of integration within existing industrial frameworks and systems. An important benefit of AutomationML is the advancing integration with the Asset Administration Shell (AAS)~\cite{AASInDetail}. AAS is a set of standards and recommendations to provide interoperability in industrial systems. AAS is typically used on the resource level (e.g., a motor or a robot). On the contrary, the AAS is used for the product level in the presented approach utilizing PoPAN. Moreover, AAS is used for the entire product life-cycle in this approach, including not only the manufacturing phase but also operations, maintenance, and repair, as well as decommissioning/recycling. The paper~\cite{Plociennik2022} highlights the use of AAS in the context of product life cycle management.

This paper addresses the following research questions:

\begin{itemize}
    \item \emph{RQ1}: How can we adapt the PAN model to primarily reflect the product structure rather than the production process with resource structure?

    \item \emph{RQ2}: Is it possible to combine production and remanufacturing/disassembling processes into a proposed PPR-based description and if yes, how to do so?

    \item \emph{RQ3}: To improve the adoption of the proposed approach by industry and academia, how can be the proposed model represented in the AutomationML data format?
\end{itemize}

The proposed modeling approach is demonstrated in a case study on disassembling electric vehicle (EV) batteries for recycling and remanufacturing~\cite{EVRecyclation}.

\section{State of the Art}

The proposed approach is built on the top of 3 main domains including (i)~Asset Administration Shell, (ii)~AutomationML data format, and (iii)~Model-Driven Engineering and Product-Process-Resource Asset Network. These research areas are described in detail in the following subsections.

\subsection{Asset Administration Shell}

One of the key standards in the Industry 4.0 initiative is the Asset Administration Shell (AAS)~\cite{AASPart1, AASPart2}. AAS was introduced by Platform Industry 4.0 as a promising approach to enable interoperability in various industries. It facilitates data sharing between value chain partners, standardizes data security, and establishes technology-neutral semantic standards, making it crucial to achieving Industry 4.0 goals. The AAS enables diverse communication channels and applications, serving as a bridge between physical objects and the digital world~\cite{AASInDetail}.

The AAS is a digital representation of an asset and comprises multiple submodels that represent a specific aspect or set of characteristics of an asset. Within a submodel, various submodel elements are defined. Submodel elements are the individual components or entities that make up the submodel. They represent specific properties, operations, parameters, references, files, or other relevant information associated with the asset. The purpose of submodels is to provide a structured approach to organizing and managing the elements within an AAS. By grouping related elements, it becomes easier to define and enforce standards for specific sets of information. 

\subsection{AutomationML}

The engineering phase of multi-disciplinary mechatronic systems requires data exchange across software tools belonging to diverse engineering domains (including electric planning, mechanic design, maintenance, etc.). Engineering data can be efficiently captured and exchanged in the data format AutomationML, which is a standardized extension of XML for the interdisciplinary exchange of planning data for production systems and processes. The data format AutomationML is an open, platform-independent standard known as IEC 62714. Detailed information about this standard including various best practices and application recommendations, and how to model specific information, can be found on the Web pages of the AutomationML association\footnote{Online: https://www.automationml.org/}.

AutomationML combines and harmonizes already existing XML formats:
\begin{itemize}
    \item  CAEX (IEC 62424) – Topology a hierarchy of objects, properties, and relationships among objects
    \item COLLADA (ISO/PAS 17506) – Geometry a kinematics of objects
    \item PLCopen XML (IEC 62714) – Discrete behavior of objects
\end{itemize}

The basis of the object model in AutomationML is CAEX, which is intended for capturing engineering information in object-oriented form. Each object can have attributes and references to other objects (i.e., internal links or mapping objects, and references via object identifiers). It can also include links to other information, represented in COLLADA or PLCopen XML. In CAEX, a model is represented via the following four basic cornerstones. Role Class Libraries are intended to define object semantics. Each object can have more than one role. Interface Class Libraries provide specifications of object interfaces. Classes of objects are modeled as objects of System Unit Class Libraries. Each system unit class poses a generic object prototype, which can be instantiated when describing a specific system. For better interoperability, semantic mapping to terminology defined as ECLASS based on IEC 61360 can be used in a standardized way. The system description itself is represented in AutomationML/CAEX by instance hierarchy. In the tree structure behind the instance hierarchy root element, Internal Elements, and their sub-elements are defined. Internal elements in most cases reference their prototypes in the frame of the System Unit Class, supported roles from Role Class Libraries, and interfaces from the interface class libraries. Topological relationships are represented as Internal Links.

AutomationML is one of the most complex data formats for system engineering, which is suitable for a wide range of tasks, including the representation of the Asset Administration Shell~\cite{Drath2019, Lueder2021}. Such an AutomationML representation can be imported to the AASX format~\cite{AASPart1} in the tool AASX Package Explorer\footnote{Online: https://github.com/eclipse-aaspe}. The application of AutomationML in the end-of-life phase is addressed in~\cite{SCHMIDT2017,Schmidt2017Indin}, which are focused on production systems and resources, whereas this paper is mainly focused on products together with production processes.

\subsection{Model-Driven Engineering and Product-Process-Resource Asset Network}

Model-driven engineering (MDE) highlights the role of models as core artifacts for engineering and integration development. Similarly, like object-orientation in object-oriented programming, where everything is represented by objects (with specific attributes and methods), in model-driven engineering, everything is a model~\cite{Bezivin2004}. Initial efforts in MDE started with the use of the Meta Object Facility and the Unified Modeling Language (UML)~\cite{Kent2002}. MDE is highly efficient for translating information between various representations (i.e., domain models or data formats)~\cite{Guerra2013, Lano2022}; and it provides suitable support for multi-disciplinary engineering projects~\cite{Berardinelli2017}. One of the plausible models to support the model-driven engineering is a Product-Process-Resource model~\cite{Meixner2020}, categorizing artifacts relevant for industrial production into these three categories including mappings or links among these categories.

The Product-Process-Resource Asset Network (PAN)~\cite{Biffl2021CBI} is a graph-based formalism for expressing relationships among products, production processes, and resources. In the domain of Cyber-Physical Production Systems (CPPSs), dependencies among products, processes, and resources are frequently left implicit. The PAN addresses this gap by providing a structured framework for explicitly articulating and overseeing the dependencies among product, process, and resource assets. The PAN modeling is suitable for capturing and sharing knowledge among engineers and respective engineering tools and for identifying changes during engineering processes of multi-disciplinary system engineering projects. PAN usage optimizes engineering processes' efficiency and enhances operations by modeling manufacturing processes, tracking resource usage, and monitoring asset performance. Through data integration, visualization, and analysis, PAN enables informed decision-making~\cite{Winkler2021ETFA}.

\section{Product-oriented Product--Process--Resource Asset Network (PoPAN)}
\label{secPoPAN}

\begin{figure*}
    \centering
    \includegraphics[width=1\linewidth]{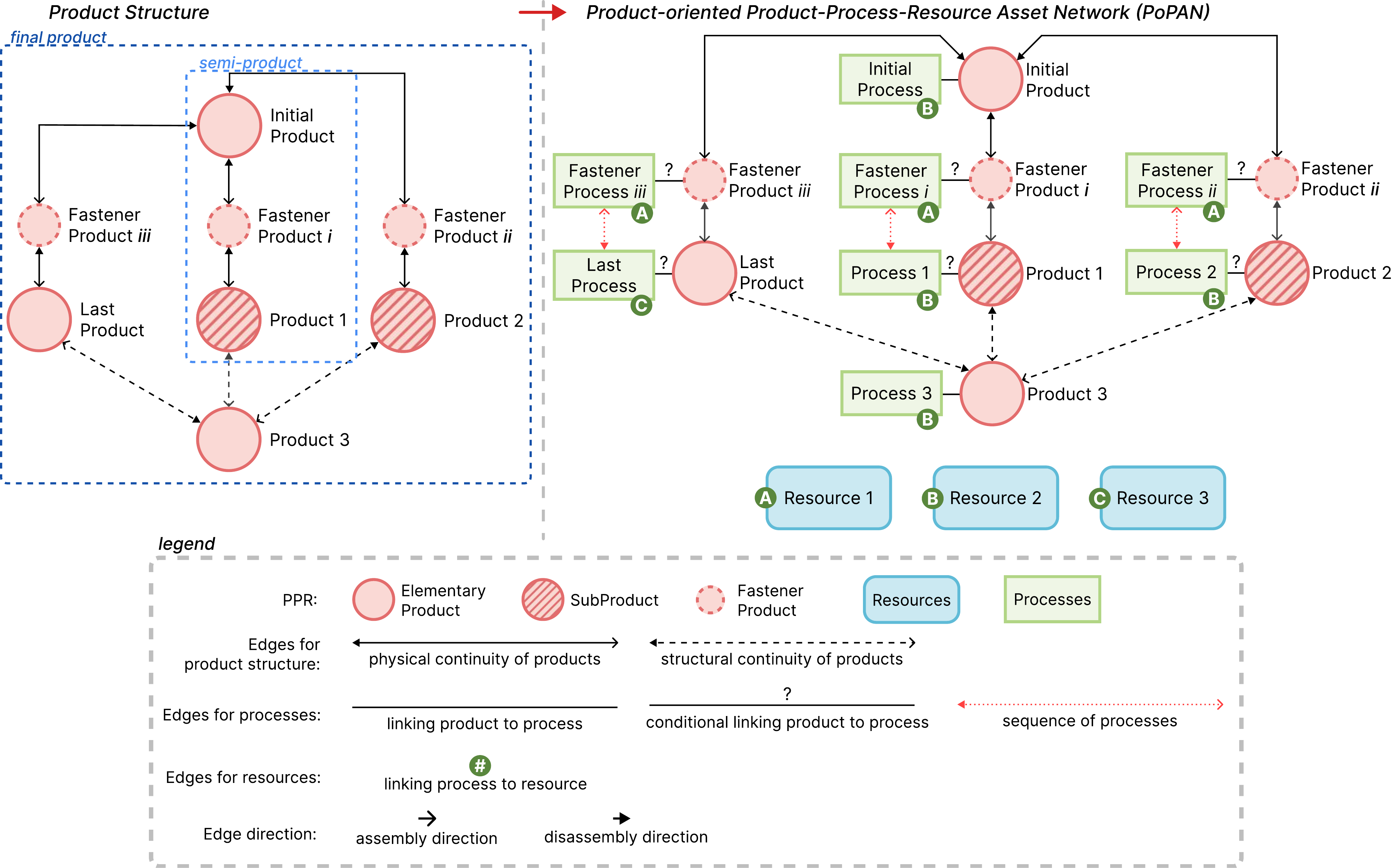}
    \caption{The proposed PoPAN model for a generic product (on the right), which has been created based on the Product Structure model (on the left). The PoPAN graph is substantiated by products, which have assigned processes, and the edges (their types, orientation, and arrow types) have a crucial role.}
    \label{figPoPAN}
\end{figure*}

Recycling and remanufacturing products pose challenges due to limited engineering data availability. While manufacturing processes are well-specified and documented, the absence of detailed data for a product in its end-of-life phase complicates recycling and remanufacturing efforts. Leveraging the PAN approach, which has proven efficiency in product manufacturing, presents a promising solution. However, traditional PAN methods describe one specific assembly process, which may not capture the entire variability of assembly and disassembly operations. To address this limitation, this paper proposes extending the PAN paradigm to describe products structurally. By doing so, multiple disassembly processes can be generated from a single description, accommodating the diverse disassembly options frequently encountered in practice. This structural representation enhances flexibility and efficiency in creating recycling and remanufacturing processes.

To create a product description that offers a comprehensive understanding of the product's structure, including the specific processes necessary for both assembly and disassembly, along with the resources required for each process, the initial step is to construct a Product Structure graph depicting dependencies between product's components (see Fig.~\ref{figPoPAN} on the left). Since the PPR approach is used, to preserve the name of elements, the product's components are depicted also as products. Within this framework, three types of products are distinguished (1) Elementary Product: a product that cannot be further disassembled, (2) SubProduct: a product that is considered elementary in terms of assembly/disassembly procedures, but is composed of several parts and can be further disassembled, and (3) Fastener Product: a product that serves as a fastener for other products (e.g., screws). The graph delineates two types of edges connecting individual products, forming the overall structure. The solid line edge illustrates physical continuity between products and the dashed line edge, on the other hand, forms a structural continuity between the products, i.e., dependencies that indicate certain products cannot be assembled or disassembled until others are. Moreover, both types of edges end with arrows on both ends representing the assembly/disassembly direction. Open arrows represent assembly direction, full arrows represent disassembly direction, and \textit{Initial Product} and \textit{Last Product} are highlighted for better orientation in the graph. For example, when disassembling \textit{Product~3}, it is essential to consider all three structural dependencies. Considering the orientation by the full arrows, \textit{Product~1} and \textit{Product~2} are not blocking \textit{Product~3} from being disassembled. However, the \textit{Last Product} is blocking the disassembly of \textit{Product~3}, meaning that the \textit{Last Product} must be disassembled before \textit{Product~3}. Additionally, frequently used terms \textit{final product} and \textit{semi-product} in the sense of a classic PAN are highlighted to show intermediate steps in an assembly operation. However, the PoPAN approach diverges from relying on these terms.

Expanding the Product Structure graph with the PAN/PPR approach involves incorporating processes and resources alongside products. This structured visualization facilitates understanding the relationships between products and corresponding processes, which are supported by the required resources. As depicted in Fig.~\ref{figPoPAN} on the right, in the extended Product Structure graph called Product-oriented Product--Process--Resource Asset Network (PoPAN), each product is connected with a corresponding process by a simple line edge or by a simple line edge with a question mark above it. The question mark indicates conditional linking between the product and process, where the process sequence might not correspond with the structure of a product, which is then resolved by a red dotted line edge determining the sequence of the processes. Lastly, the graph contains green circle marks with a letter which depict edges connecting processes with their required resources. As with the previous Product Structure graph, the arrows form the sense of orientation in the graph and the connections between products, processes, and resources form the edges.

To find the correct sequence of processes for the assembly/disassembly operation within the PoPAN, it is necessary to follow a set of certain rules. In addition to the assembly/disassembly direction of orientation according to the arrows, special attention must be paid to the priority of the edges. For example, the steps taken when following the assembly direction from \textit{Initial Product} to the \textit{Fastener Product i} and \textit{Product 1} (Fig.~\ref{figPoPAN}) are as follows:

\begin{enumerate}
    \item Upon arriving at \textit{Fastener Product i}, the conditional linking product to the process edge is examined.
    \item Continuing to \textit{Fastener Process i}, the sequence of processes edge is checked, revealing a discrepancy in direction.
    \item Returning to \textit{Fastener Product i}, the physical continuity of the product's edge is verified in the correct direction, leading to \textit{Product 1}.
    \item Upon reaching \textit{Product 1}, the structural continuity of the product's edge is examined, indicating the \textit{Product 3} is to be assembled subsequent to \textit{Product 1}.
    \item Next, the conditional linking product to process edge is followed, leading to \textit{Process 1}.
    \item Verifying the edge sequence of processes at \textit{Process 1}, it's confirmed to be in the right direction.
    \item \textit{Process 1} with \textit{Product 1} is noted simultaneously with \textit{Resource 2}, and the sequence of processes edge is followed back to \textit{Fastener Process i}.
    \item \textit{Fastener Process i} is noted with \textit{Fastener Product i} simultaneously with \textit{Resource 1}.
\end{enumerate}

As outlined in the preceding steps, when assembling a product the initial edge to examine, if present, is the structural continuity of the product's edge. This edge determines whether the product can be assembled before other products. Once the feasibility of assembly/disassembly is established, the next edges to consider are the linking products to the process edge and conditional linking products to the process edge, which assign processes to products. Additionally, for the conditional linking products to the process edge, the sequence of the process edge is presented which ensures the processes are sequenced correctly. Afterward, the linking product to the process edge is followed to assign a resource to a product after which the physical continuity of the product's edge can be taken to continue to the next product where the whole process is repeated.

These rules, which specify the order in which edges should be traversed along with the assembly/disassembly direction, enable PoPAN to determine the correct path for creating assembly/disassembly operations. This capability facilitates the efficient design of recycling or remanufacturing processes of products.

\section{Representation of PoPAN in AutomationML}
\label{secPoPANAML}

\begin{figure}
    \centering
    \includegraphics[width=0.45\linewidth]{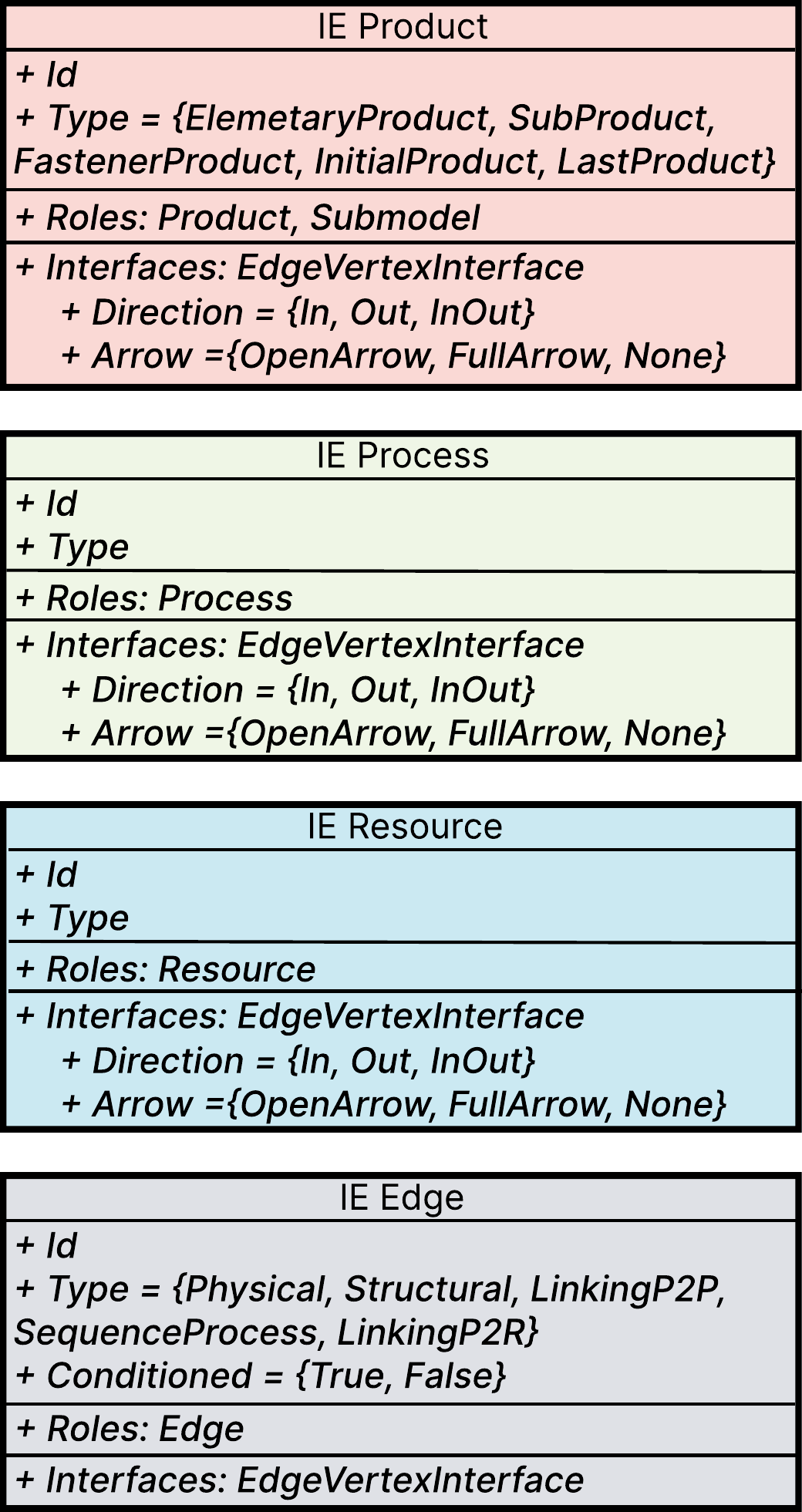}
    \caption{System Unit Classes of the AutomationML description for PoPAN representation.}
    \label{figAMLSUC}
\end{figure}

\begin{figure}[t]
    \centering
    \includegraphics[width=0.55\linewidth]{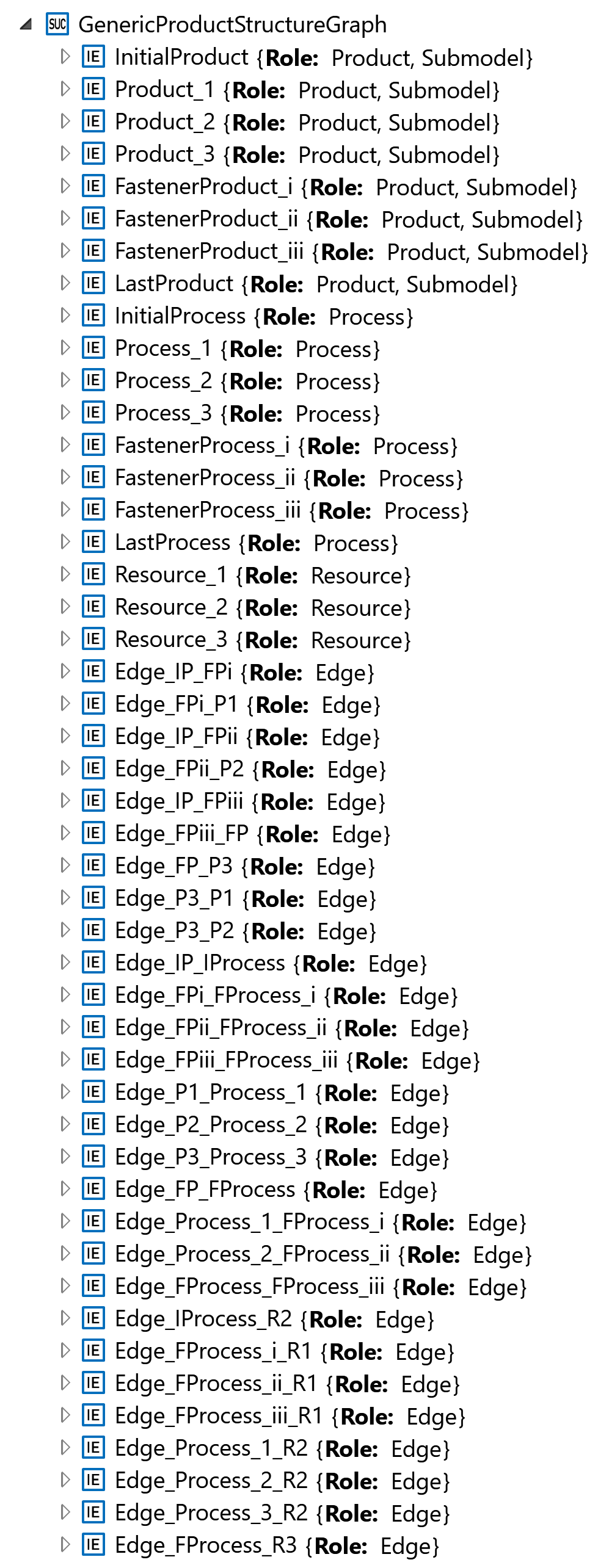}
    \caption{Example of a generic product structure in the AutomationML Editor.}
    \label{figAMLGenericProductStructureAMLEditor}
\end{figure}

\begin{figure*}
    \centering
    \includegraphics[width=1\linewidth]{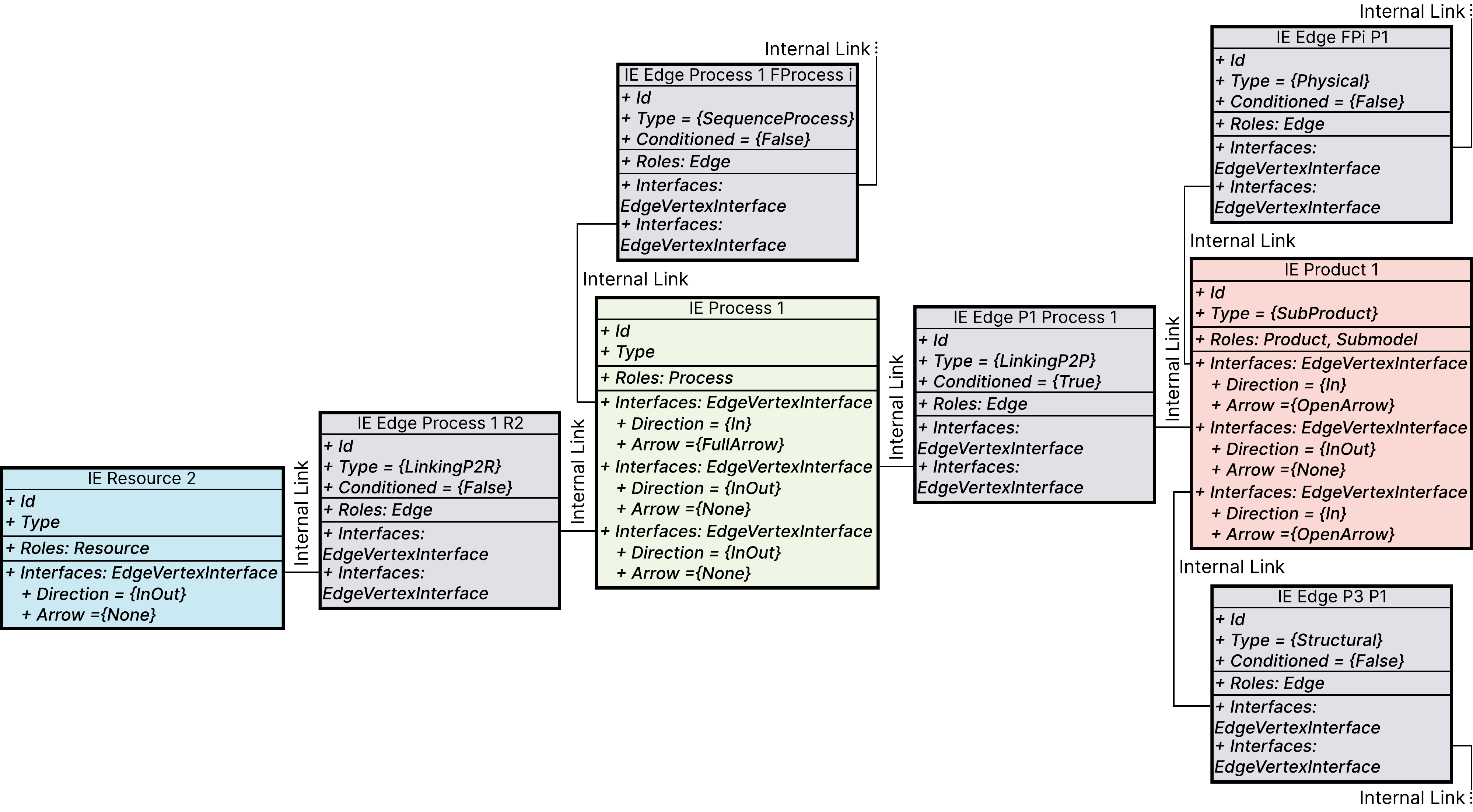}
    \caption{Example of a specific PoPAN represented in the AutomationML data format. It includes Internal Elements (IE) for Resources (blue), Edges of PoPAN (grey), Process (green), and Products (salmon).}
    \label{figAML}
\end{figure*}

Having the PoPAN model, it is necessary to serialize such a graph into a computer-understandable form to capture the knowledge and support its exchange across involved stakeholders. The following requirements on the data format were postulated: (i) widely adopted and standardized data format, (ii) support for AAS, (iii) modularity and scalability of the description, and (iv) platform neutrality and openness. Considering these requirements, we decided to use the AutomationML data format for serialization of the PoPAN description.

The AutomationML representation aims at capturing the PoPAN network as a graph. First, we specified the InterfaceClassLib with the Interface Class \textit{EdgeVertexInterface}. This interface allows the connection of edges and vertices via AutomationML Internal Links. The Interface Class has an attribute \textit{Direction}, which can have values \textit{In} for the incoming edge, \textit{Out} for the outgoing edge, and \textit{InOut} for the undirected edge. Since PoPAN distinguishes assembly and disassembly directions, the \textit{EdgeVertexInterface} Class has also attribute \textit{Arrow}, which can have values \textit{OpenArrow} for the assembly direction, \textit{FullArrow} for the disassembly direction, or \textit{None} for none arrow.

In the second step, we have defined the Role Class Library with the following terms for PoPAN specifications: (i)~\textit{Product}, (ii)~\textit{Process}, (iii)~\textit{Resource}, and (iv)~\textit{Edge}. All these Role Classes have attributes \textit{Id} and \textit{Type} (to unambiguously specify these assets) and one or more \textit{EdgeVertexInterface} (to enable connecting edges via Internal Links).

In the third step, we have defined exemplary/generic System Unit Class, which are depicted in Fig.~\ref{figAMLSUC}. The figure shows four Internal Elements (IE) from the System Unit Class, in compliance with their Roles from the Role Class Library. The Internal Elements are as a matter of fact instances of the aforementioned Role Classes, accompanied by \textit{EdgeVertexInterface} to enable the PoPAN representation.

The IE Product type represents physical products within the whole product. Each IE Product is uniquely identified by an \textit{Id} and has a \textit{Type} including \textit{ElementaryProduct}, \textit{SubProduct}, \textit{FastenerProduct}, \textit{InitialProduct}, or \textit{LastProduct}. Additionally, IE Product has assigned Roles \textit{Product} and \textit{Submodel} (to correspond with AAS).
The IE Process type represents processes performed within the assembly/disassembly operations. IE Process is uniquely identified by an \textit{Id}, has assigned Role \textit{Process}, and has a \textit{Type}.
The IE Resource type represents resources needed to perform processes, such as tools and stations. Similar to IE Process, each IE Resource has a unique identifier \textit{Id}, has an assigned Role \textit{Process}, and has a \textit{Type}.
The IE Edge type represents connections between PPR components within the whole product. IE Edge is uniquely identified by \textit{Id}, the Role \textit{Edge} and has a \textit{Type} that can have values \textit{Physical} to represent the physical continuity of products, \textit{Structural} for the structural continuity of products, \textit{LinkingP2P} for linking a product to a process, \textit{SequenceProcess} to represent dependencies/sequences of processes, and \textit{LinkingP2R} for linking a process to a resource.
For all four IE Product types, the \textit{EdgeVertexInterface} is defined to ensure connection to other IEs within the PoPAN.

The IE Product supports the Role \textit{Submodel} from the \textit{Asset Administration Shell Role Class Library}\footnote{Online: https://www.automationml.org/wp-content/uploads/2022/04/Asset-Administration-Shell-Representation-V1\_0\_0.zip}. In this place we should note that we are missing a ``SubModelElement/Entity'' in this AutomationML library, which would be useful for transforming the description into the AASX format.

The example of a generic product structure according to the PoPAN model from the previous section is depicted in Fig.~\ref{figAMLGenericProductStructureAMLEditor}. If we need to assign more than one edge to a vertice, we have to increase the number of \textit{EdgeVertexInterfaces}. In Fig.~\ref{figAML}, a network of PoPAN elements represented in AutomationML is shown for a \textit{Product 1} from a generic product (Sec.~\ref{secPoPAN}).

AutomationML supports annotating objects relevant to the Asset Administration Shell (AAS). AAS is becoming an important industrial artifact to be exchanged in the entire supply chain. For creating the AAS, we are using the aforementioned tool AASX~Package Explorer, which can import an AutomationML file and transform it into the AASX format, which can accompany the product during its entire life-cycle.

This comprehensive product description, when used in conjunction with the AAS, can accompany the product throughout its entire life-cycle. It can assist in scenarios where defective parts need replacement, while also facilitating the tracking of new or missing products (e.g., a missing screw) to streamline the recycling process.

\section{Electric Vehicle Battery Remanufacturing Use-Case}

To illustrate the advantages of the PoPAN approach and to test its validity and benefits for industrial applications, electric vehicle (EV) battery remanufacturing serves as a use-case. From a circular economy perspective, where resources are reused to minimize waste and enhance sustainability, EV battery recycling/remanufacturing is a significant concern. This is due to the valuable materials contained within EV batteries and the environmental risks posed by improper management of EV batteries in their end-of-life phase.

\begin{figure}
    \centering
    \includegraphics[width=1\linewidth]{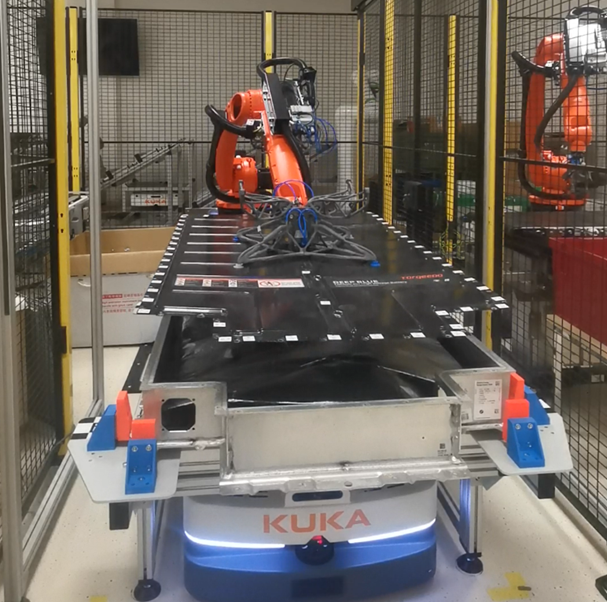}
    \caption{A KUKA robot is lifting the lid of an EV battery from a BMW i3 in the Industry 4.0 Testbed at CIIRC}
    \label{figBatteryTestbed}
\end{figure}

\begin{figure*}
    \centering
    \includegraphics[width=1\linewidth]{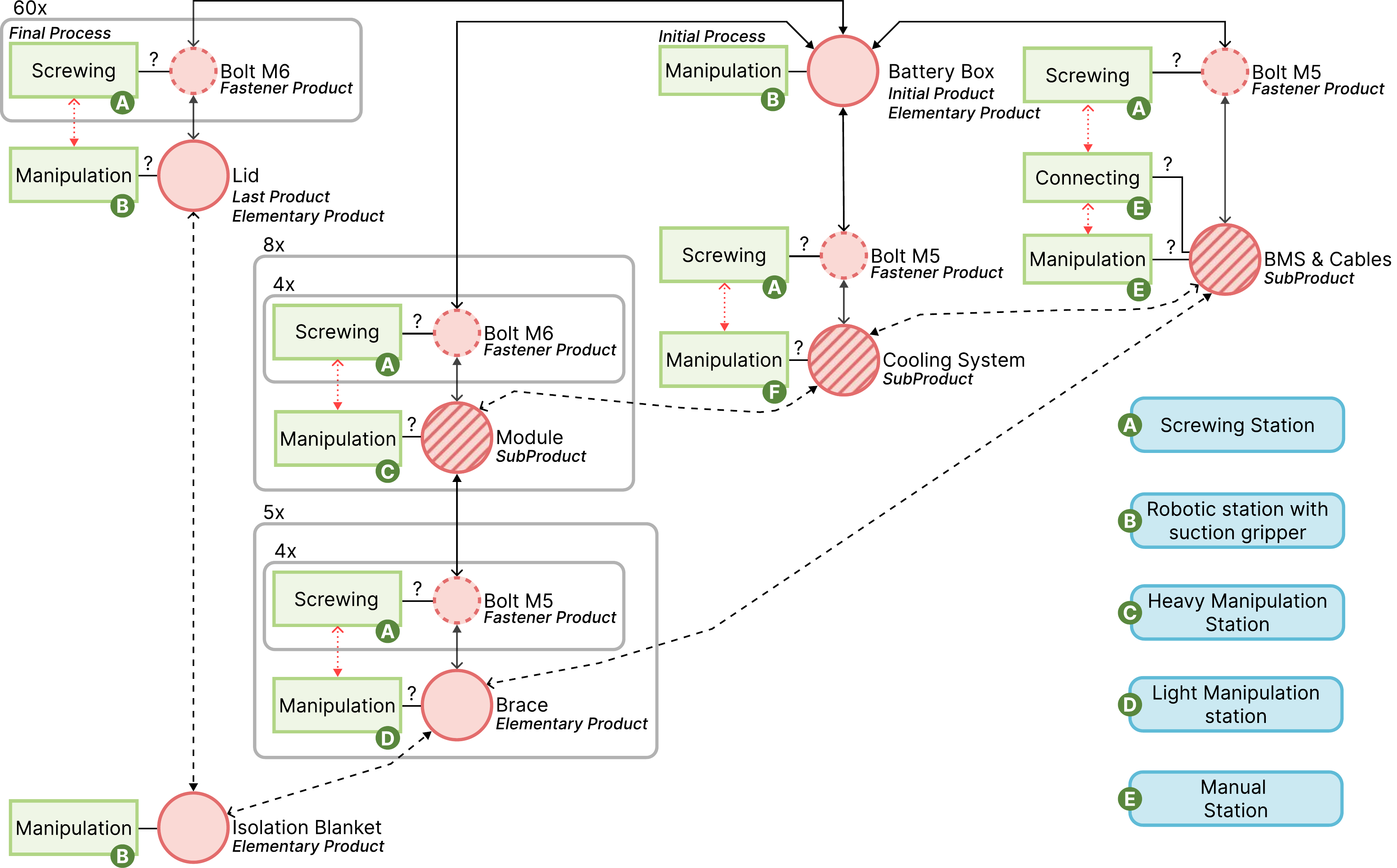}
    \caption{PoPAN describing a simplified structure of the EV battery from BMW i3. Products are accompanied by production processes and resources, which enables to find assembling and disassembling operations from one graph.}
    \label{figPANBat}
\end{figure*}

In our laboratory conditions, utilizing the Industry 4.0 Testbed\footnote{Online: https://www.ciirc.cvut.cz/teams-labs/testbed/} at CTU in Prague -- CIIRC, we employ KUKA robots to manipulate the real EV battery according to predefined parameters (see Fig.~\ref{figBatteryTestbed}). 
The primary goal of this use-case is to disassemble the entire EV battery with a flexible robotic system. This approach offers the potential use of modules from the EV battery for secondary applications. Additionally, the secondary goal involves replacing broken modules, which necessitates a combination of assembling and disassembling operations.
We were seeking a formalism that allows to accommodate all possible operations in our use-case, and PoPAN fits our requirements perfectly for all considered applications. The visualization of PoPAN for a simplified description of the EV battery is depicted in Fig.~\ref{figPANBat}. It illustrates all relevant data for creating assembly/disassembly operations. As outlined in Section~\ref{secPoPAN}, this visualization contains products, processes, and resources interconnected by different types of edges, with two types of arrows indicating assembly/disassembly orientations. Following the rules described in Section~\ref{secPoPAN}, remanufacturing and recycling operations can be effectively designed. Through integration with the AAS, PoPAN serves as a digital shadow for the EV battery, accompanying it throughout the entire life-cycle. Such an integration enhances data management and data sharing between involved stakeholders.

From the PoPAN presented in Fig.~\ref{figPANBat}, we can query the set of operations to disassemble the battery in the robotic workcells.
We can see that the \textit{Last Product} is the Lid, which has to be manipulated/removed from the EV battery. However, the conditioned edge between the  Lid product and Manipulation in this PoPAN requires to evaluate the edges connected to the vertex Manipulation. Since we are currently disassembling, we cannot proceed to the process Screwing. The search algorithm proceeds in the full arrow direction to the Bolts M6 with the assigned Screwing process. When finished, it can continue to the Manipulation process, which was previously skipped. Then the search algorithm can proceed to the product Battery Box, which has more than one incoming edge. Prior to executing the assigned process Manipulation, the incoming edges have to be passed. Therefore, the search algorithm proceeds to the remaining branches and starts from the vertex that has all dependencies satisfied. During this systematic search, it goes through the entire graph until all processes are done to get to the \textit{Initial Product}, which is the empty battery box in this case.

Through the performed evaluation, we have confirmed the efficacy of PoPAN, as it seamlessly integrates all tasks developed in this use-case. Based on this lessons-learned use-case, we believe PoPAN can be also well-suited for other use-cases.

\section{Conclusion and Future Work}

This paper proposes a specific adaptation of the PAN paradigm called Product-oriented PPR Asset Network (PoPAN), having the product structure as the core basis of the descriptive model. Such an adaptation enables to express not only the assembling processes but also the disassembling ones in just one compact form. The PoPAN model can be afterward queried to get the right sequence of production processes/operations for various production, reparation, remanufacturing, or recycling purposes during later phases of the whole product life-cycle.

The PoPAN model serves as a model-driven engineering backbone for process management within the whole product life-cycle.
It enables large-scale system integration and scalability of the models and respective systems. The proposed approach is demonstrated in a case study on disassembling EV batteries to upcycle them from automotive applications towards stationary battery storages in industry or households.

Addressing the research question \emph{RQ1}, this paper proposes a PPR-based model coming up from the PAN model, which however primarily reflects the product structure. The model allows to seamlessly combine the production and remanufacturing/disassembling processes into one unified description, addressing the research question \emph{RQ2}. This overall model was presented in Sec.~\ref{secPoPAN}.

During the presented research, we were aware of the necessity to foster the adaptability of this new modeling approach by industrial and academic stakeholders. Therefore, we have from the very beginning considered an export to the AutomationML data format and Asset Administration Shell, addressing the research question \emph{RQ3}, providing the serialization of the model into AutomationML in Sec.~\ref{secPoPANAML}.

In future work, we would like to implement a user-friendly search algorithm for generating the right directed acyclic graph of processes for the queried situation and export it into BPMN.

\section*{Acknowledgment}
This work was supported by the Grant Agency of the Czech Technical University in Prague, grant No. SGS24/125/OHK2/3T/12
and the project “Regeneration of used batteries from Electric Vehicles” (Slovak ITMS2014+ code 313012BUN5), the project is part of the Important Project of Common European Interest (IPCEI) called the European Battery Innovation (code OPII-MH/DP/2021/9.5-34), which was announced as a part of Operational Program Integrated Infrastructure (EZOP ID 71235). This work was co-funded by the European Union under the project Robotics and advanced industrial production -- ROBOPROX (reg. no. CZ.02.01.01/00/22\_008/0004590).

\bibliographystyle{IEEEtran}
\bibliography{references}

\end{document}